\documentclass[letterpaper, 10pt, conference]{ieeeconf} 

\IEEEoverridecommandlockouts           

\makeatletter
\let\NAT@parse\undefined
\makeatother

\usepackage[utf8]{inputenc} 
\usepackage{amsmath} 
\usepackage{amssymb}  
\usepackage{graphicx}
\usepackage{grffile}
\usepackage{caption}
\usepackage{subcaption}
\usepackage{tabularx}
\usepackage[hyphens]{url}
\usepackage{cite}
\usepackage{tikz}
\usepackage[mode=buildnew]{standalone}
\usetikzlibrary{calc,shapes,arrows,positioning,fit}
\usetikzlibrary{intersections}
\usetikzlibrary{backgrounds}

\usepackage{ifthen}
\newboolean{include-notes}

\setboolean{include-notes}{true}
 \newcommand{\kknote}[1]{\ifthenelse{\boolean{include-notes}}%
 {\textcolor{orange}{\textbf{K: #1}}}{}}
 \newcommand{\tbnote}[1]{\ifthenelse{\boolean{include-notes}}%
 {\textcolor{cyan}{\textbf{Tapo: #1}}}{}}
 \newcommand{\ssnote}[1]{\ifthenelse{\boolean{include-notes}}%
 {\textcolor{red}{\textbf{SS: #1}}}{}}
 \newcommand{\aknote}[1]{\ifthenelse{\boolean{include-notes}}   %
 {\textcolor{magenta}{\textbf{AK: #1}}}{}}
  \newcommand{\ahnote}[1]{\ifthenelse{\boolean{include-notes}}%
 {\textcolor{blue}{\textbf{K: #1}}}{}}
 \newcommand{\glnote}[1]{\ifthenelse{\boolean{include-notes}}%
 {\textcolor{brown}{\textbf{GL: #1}}}{}}
 \newcommand{\avknote}[1]{\ifthenelse{\boolean{include-notes}}%
 {\textcolor{green}{\textbf{AVK: #1}}}{}}
 \newcommand{\sknote}[1]{\ifthenelse{\boolean{include-notes}}%
 {\textcolor{red}{\textbf{SK: #1}}}{}}

\usepackage{hyperref}
\hypersetup{bookmarks=true,bookmarksopen,bookmarksnumbered,
pdfpagemode=UseOutlines,
breaklinks=true,
colorlinks=true,
linkcolor=blue,
anchorcolor=blue,
citecolor=blue,
filecolor=blue,
menucolor=blue,
urlcolor=blue
}
 
\usepackage{setspace}
     
\title{\LARGE \bf
Telemanipulation with Chopsticks: Analyzing Human Factors in User Demonstrations
}

\author{Liyiming Ke$^{1}$, Ajinkya Kamat$^{1}$, Jingqiang Wang$^{2}$,\\ Tapomayukh Bhattacharjee$^{1}$, Christoforos Mavrogiannis$^{1}$, Siddhartha S. Srinivasa$^{1}$
\thanks{$^{1}$Paul G. Allen School of Computer Science and Engineering,
University of Washington, Seattle, WA 98105, USA
{\tt\small \{kayke,arkamat,tapo,cmavro,siddh\}@cs.uw.edu}}%
\thanks{$^{2}$Department of Mechanical Engineering, University of Washington, Seattle, WA 98105, USA {\tt\small jwq123@uw.edu}}%
}

\begin{document}
\maketitle
\thispagestyle{empty}
\pagestyle{empty}


\begin{abstract}

Chopsticks constitute a simple yet versatile tool that humans have used for thousands of years to perform a variety of challenging tasks ranging from food manipulation to surgery. Applying such a simple tool in a diverse repertoire of scenarios requires significant adaptability. Towards developing autonomous manipulators with comparable adaptability to humans, we study chopsticks-based manipulation to gain insights into human manipulation strategies. We conduct a within-subjects user study with 25 participants, evaluating three different data-collection methods: normal chopsticks, motion-captured chopsticks, and a novel chopstick telemanipulation interface. We analyze factors governing human performance across a variety of challenging chopstick-based grasping tasks. Although participants rated teleoperation as the least comfortable and most difficult-to-use method, teleoperation enabled users to achieve the highest success rates on three out of five objects considered. Further, we notice that subjects quickly learned and adapted to the teleoperation interface. Finally, while motion-captured chopsticks could provide a better reflection of how humans use chopsticks, the teleoperation interface can produce quality on-hardware demonstrations from which the robot can directly learn.

\end{abstract}


\section{Introduction}


Roboticists have used both complex tools (such as universal grippers~\cite{Brown18809}, vacuum suction tools~\cite{Robertsoneaan6357}, and anthropomorphic hands \cite{kontoudis15, xu16}) and simple ones (such as parallel-jaw grippers and forks~\cite{monkman2007robot, mason2011generality, rodriguez2014manipulation,zisimatos14,odhner14,bhattacharjee2019towards, rhodes2018robot, melchiorri2016robot, yamazaki2010recognition}) for manipulation. Complex tools can be customized for specialized manipulation tasks, while simple tools require adaptive manipulation strategies for different usage scenarios. Studying the adaptability humans demonstrate in using simple tools could help us extract insights for developing autonomous manipulators with comparable flexibility to humans. 

\begin{figure}[t!]
\vspace{.5em}
\centering
\begin{subfigure}{.23\textwidth}
  \centering
  \includegraphics[width=\linewidth]{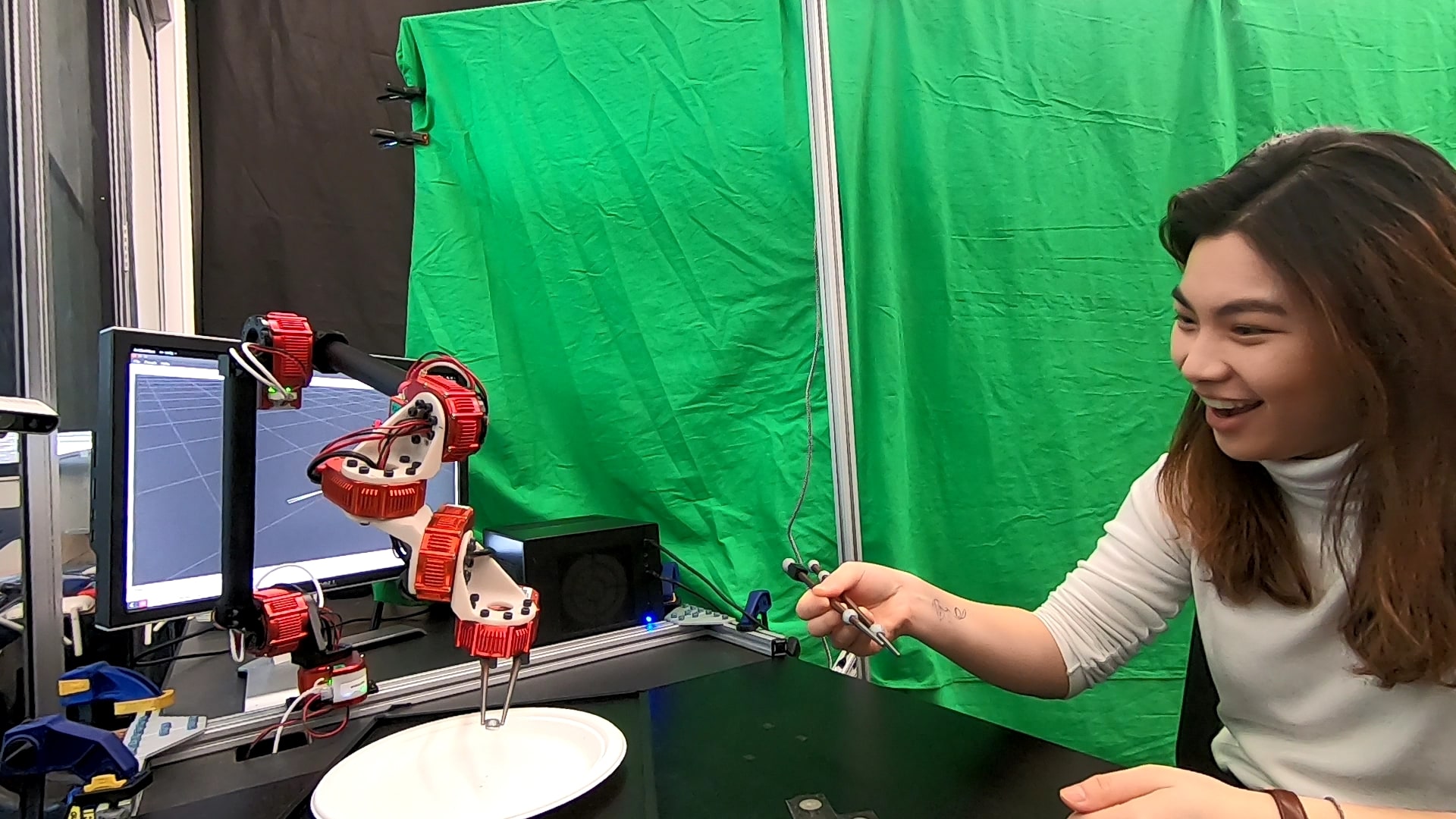}
  \label{fig:403}
\end{subfigure}
\begin{subfigure}{.23\textwidth}
  \centering
  \includegraphics[width=\linewidth]{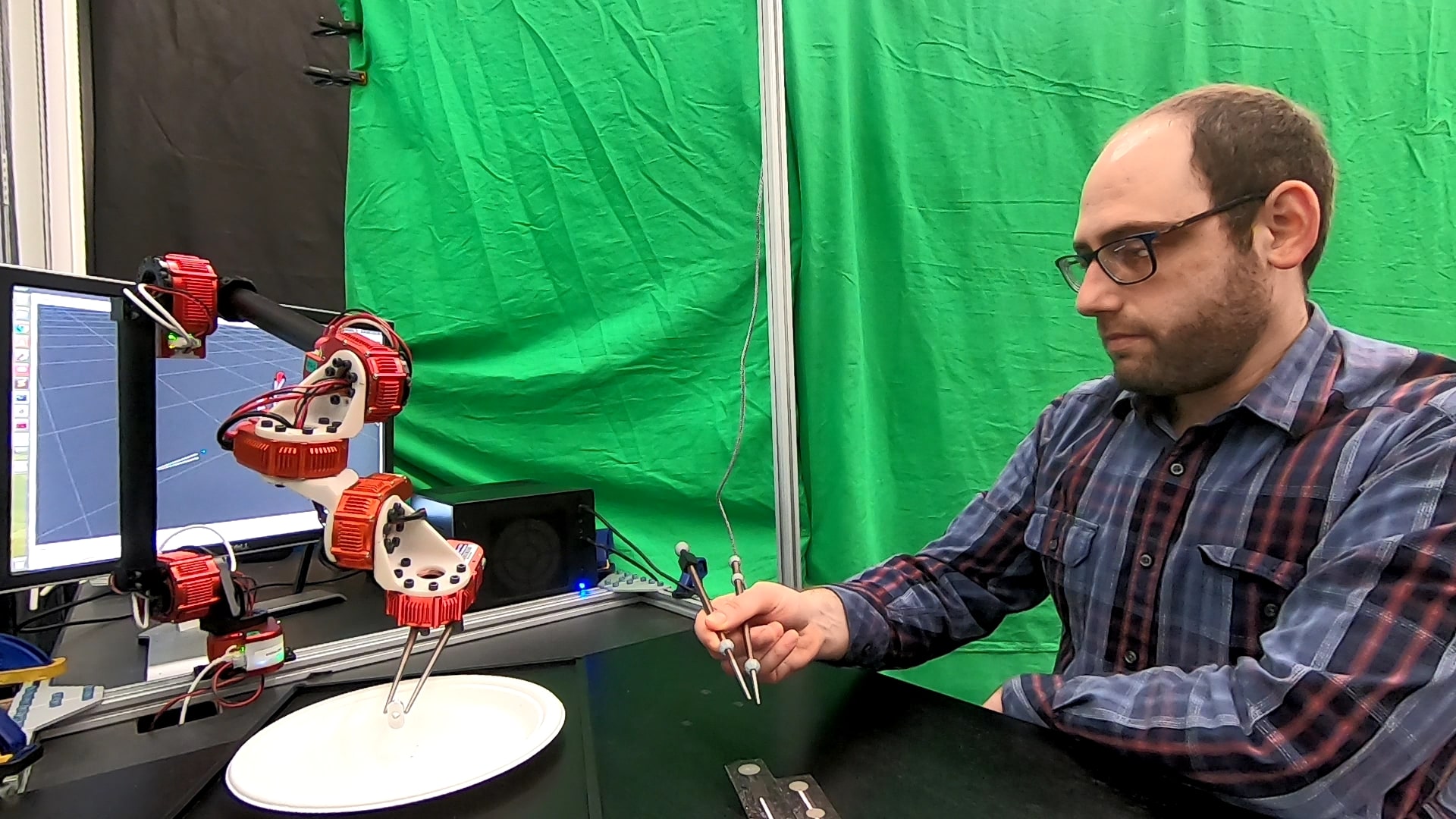}
  \label{fig:405}
\end{subfigure}
\begin{subfigure}{.23\textwidth}
  \centering
  \includegraphics[width=\linewidth]{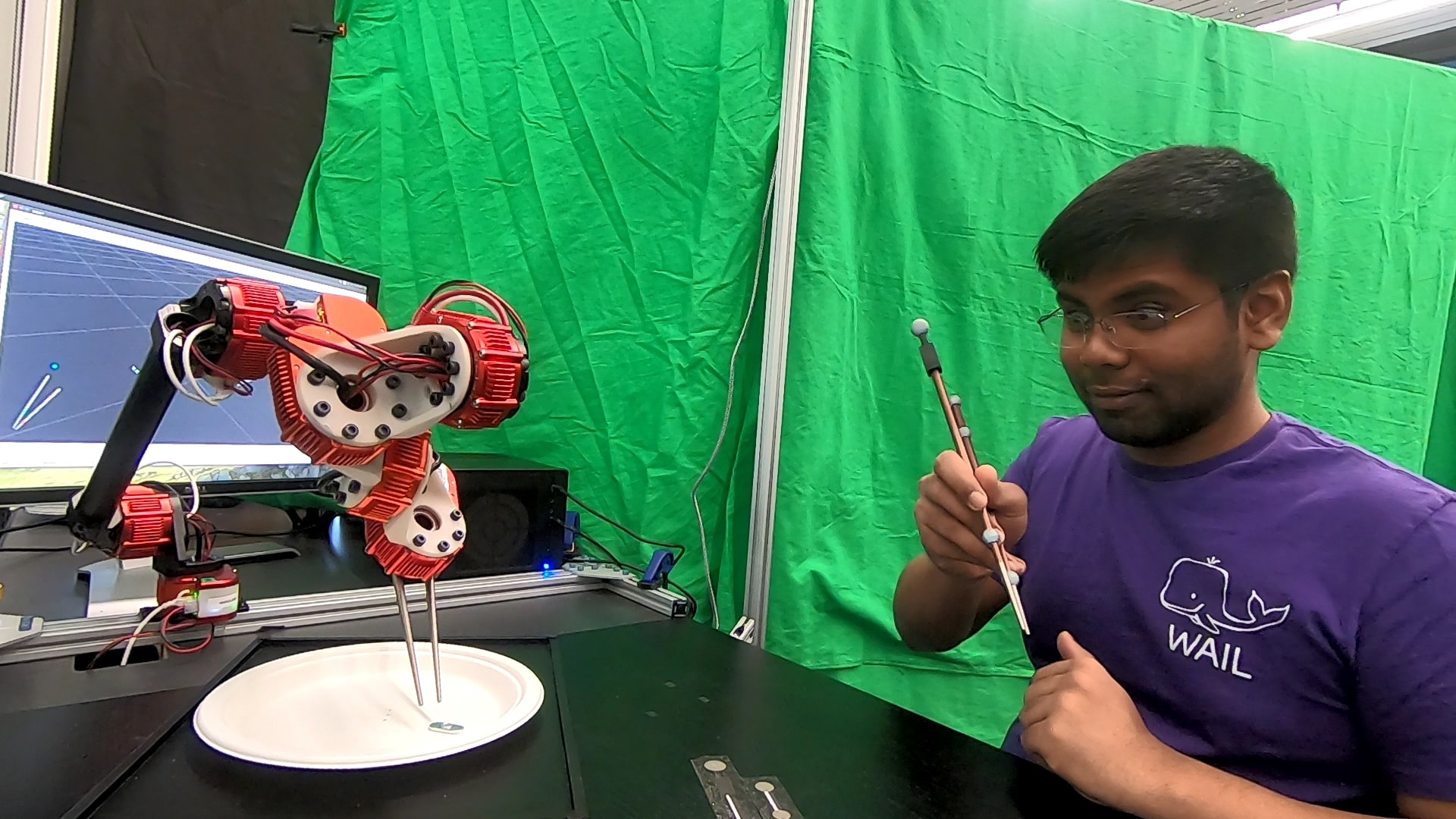}
  \label{fig:303}
\end{subfigure}
\begin{subfigure}{.23\textwidth}
  \centering
  \includegraphics[width=\linewidth]{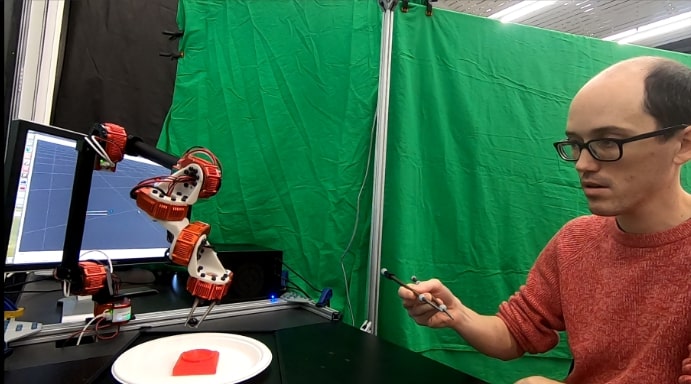}
  \label{fig:302}
\end{subfigure}
\caption{Subjects teleoperating a robot holding chopsticks using a motion-capture-marker instrumented chopstick to pick up different objects.\label{fig:subjects}}
\vspace{-.5em}
\end{figure}

In this paper, we focus on chopsticks, a simple tool that many humans are already familiar with. Every day, billions of people use chopsticks for food-related manipulation, i.e., to pick up sushi, fried rice, or twirl noodles. One wily and dexterous human even used chopsticks to pick-pocket cellphones~\cite{choppick}. Researchers have adopted the general design of chopsticks for various applications, such as meal assistance~\cite{chang2007pincer, yamazaki2012autonomous}, surgery~\cite{sakurai2016thin}, micro-manipulation~\cite{ramadan2009developmental}, repetitive pick-and-place~\cite{vassileva2009learning, pastor2011skill} and articulated-hand design ~\cite{sugiuchi2002execution, wang1988chopstick, chepisheva2016biologically}. A chopsticks-wielding robot has a potentially versatile use case to pick up objects of diverse shape, size, weight distribution, and deformation. The design of chopsticks can be easily extended and its versatility may apply to other robots, e.g., a surgical robot with a chopsticks-shaped end effector may effortlessly switch between grasping, moving and rotating various shapes of tissues with different deformability without switching hardware~\cite{sakurai2016thin}. 

\begin{figure*}[t!]
\vspace{1em}
\centering
\begin{subfigure}{.43\textwidth}
  \centering
  \includegraphics[height=3.8cm]{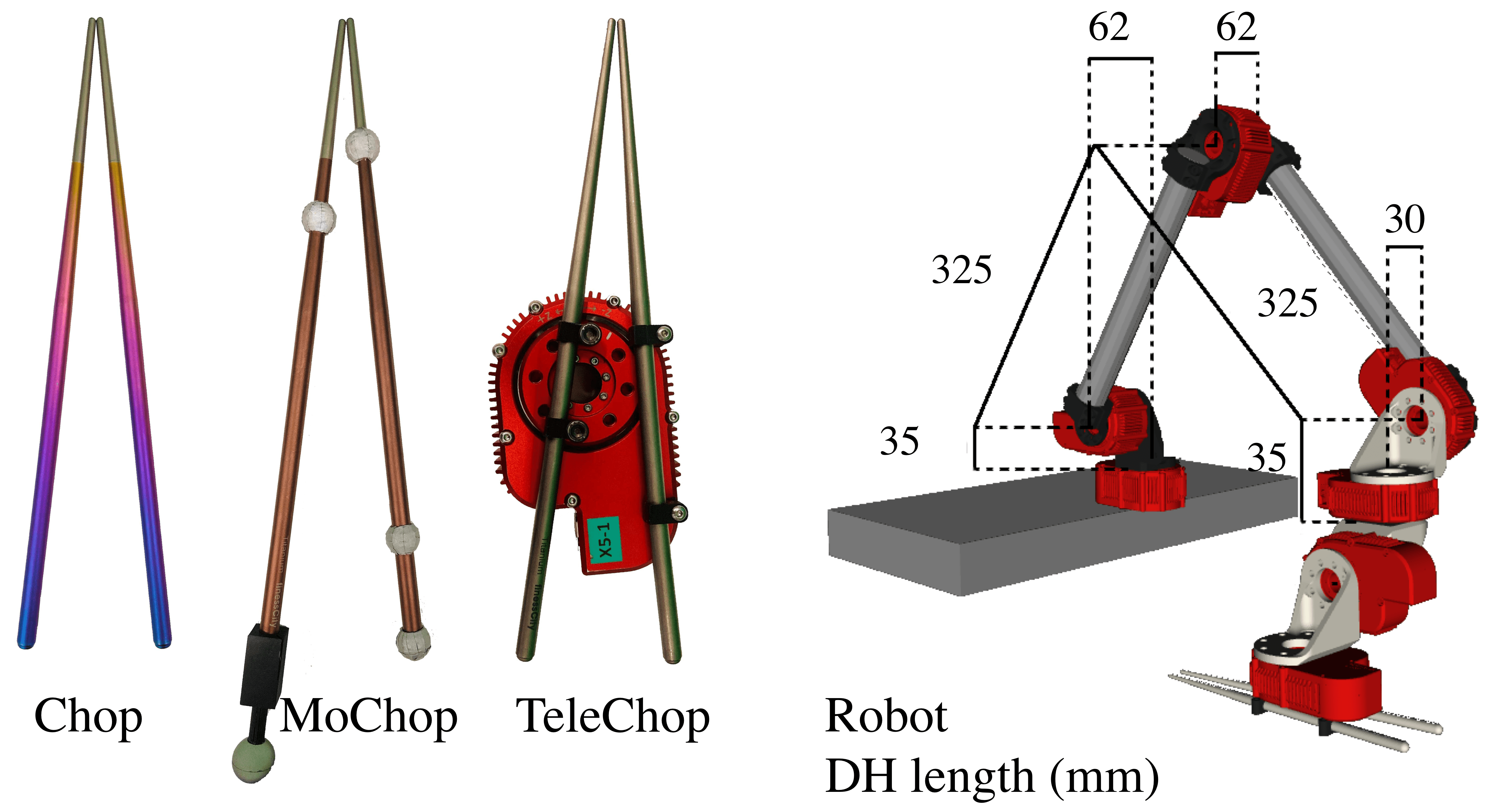}
  \caption{Hardware}
  \label{fig:system_hardware}
\end{subfigure}
\begin{subfigure}{.32\textwidth}
  \centering
  \includegraphics[height=3.8cm]{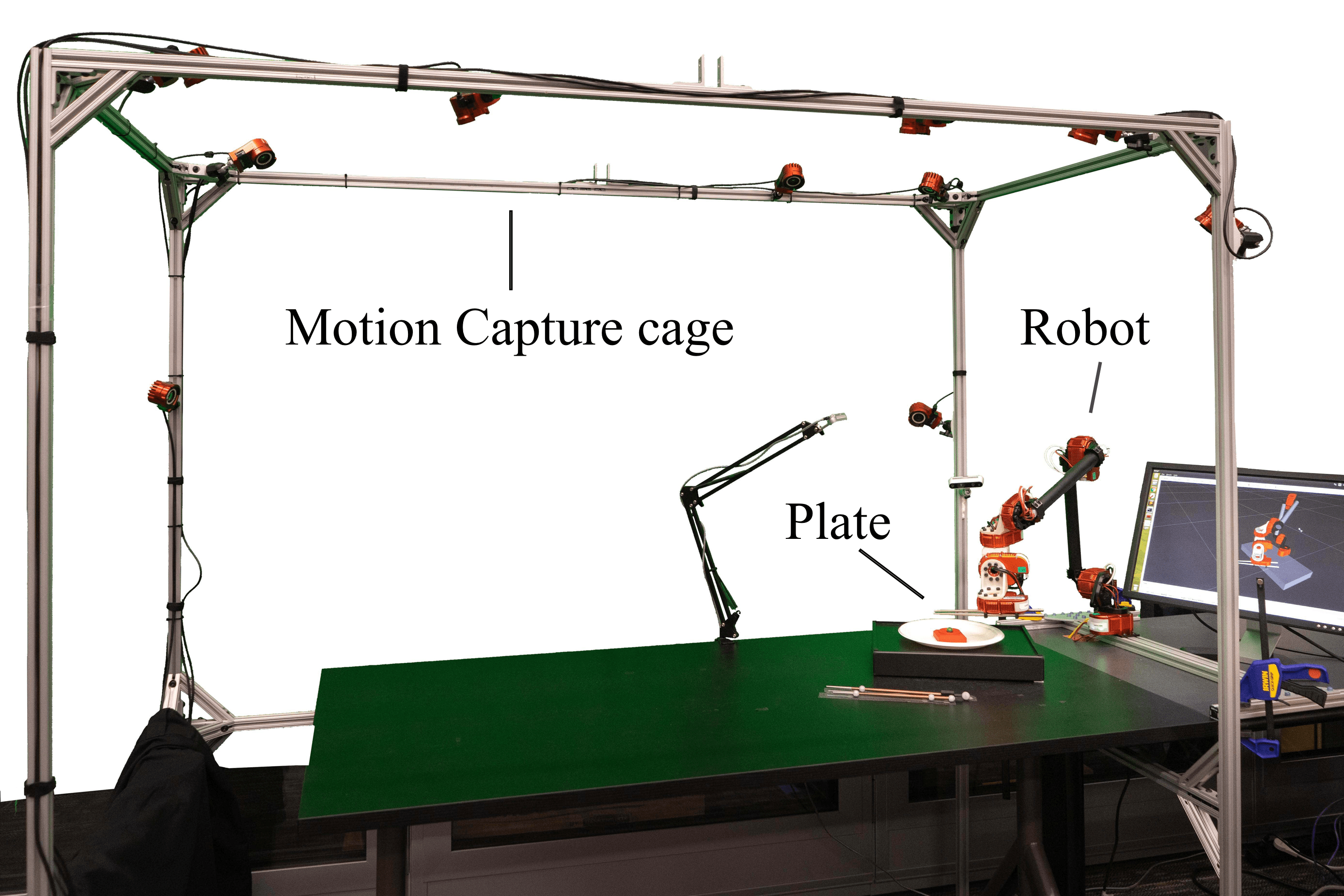}
  \caption{Motion capture}
  \label{fig:system_mocapcage}
\end{subfigure}
\begin{subfigure}{.22\textwidth}
  \centering
  \includegraphics[height=3.8   cm]{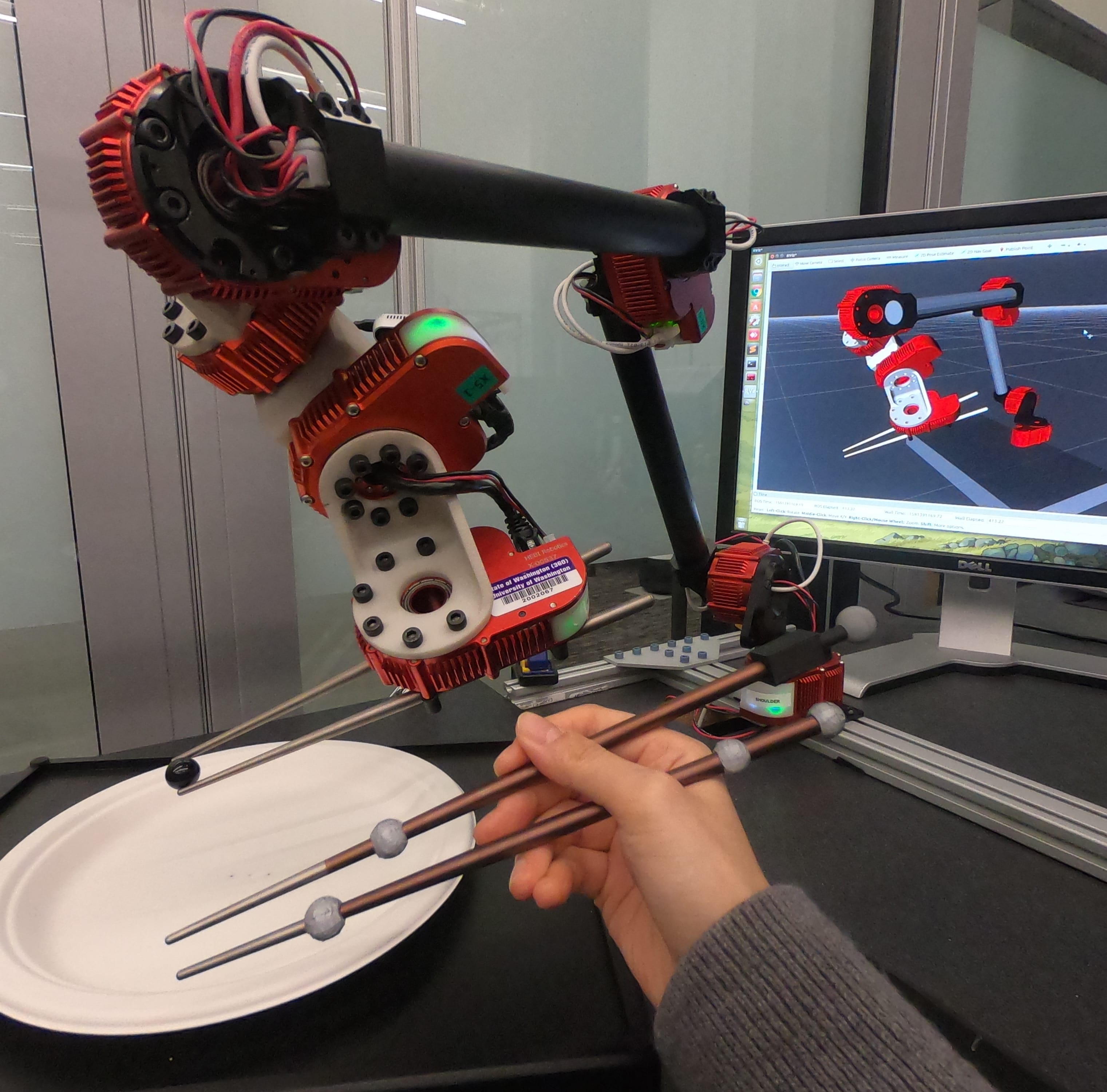}
  \caption{Teleoperation in action}
  \label{fig:system_teleop}
\end{subfigure}
\caption{Overview of our system\label{fig:system}. }
\end{figure*}

Importantly, human familiarity with chopsticks opens up the possibility of \emph{easily collecting expert demonstrations}. Despite the wide range of applications involving chopsticks, we are not aware of any prior efforts to learn from human demonstrations for chopstick-based robot manipulation tasks. To this end, we analyze how different interfaces and user expertise levels affect the quality of demonstrations by comparing three data-collection methods (Fig. \ref{fig:system_hardware}): normal chopsticks (“Chop”), motion captured chopsticks (“MoChop”), and a teleoperation interface (“TeleChop”). We conduct a within-subjects user study with 25 participants and examine how human factors affect the success rate of picking up everyday-life objects.

One key finding is that user preferences for input modalities may not necessarily correlate to the quality of the resulting demonstrations. Although users rated teleoperation as unnatural and uncomfortable, they exhibited comparable performance using TeleChop, MoChop, and Chop. Surprisingly, TeleChop enabled users to score the highest success rates on picking up three of five objects even though it lacks haptic feedback. Furthermore, since teleoperation moves the robot to complete the task, it provides directly usable demonstrations for data-driven techniques such as imitation learning. 

Overall, we make the following contributions:
\begin{enumerate}
\item An analysis of human factors underlying user demonstrations in chopstick-based telemanipulation.
\item A comparative evaluation of three different technologies for chopstick-based manipulation across a series of grasping tasks of varying difficulty: normal chopsticks, motion-captured chopsticks, and teleoperated chopsticks. 
\item A novel teleoperation interface featuring a custom-built, inexpensive 6-DOF robot manipulator, equipped with a pair of chopsticks attached to its end-effector and the ability to partially filter vibrations arising from users' grips or motion tracking.
\end{enumerate}

Humans successfully teleoperated our robot to complete a challenging manipulation task: picking up a slippery glass ball with slippery metal chopsticks, without haptic feedback. We believe that teleoperation could be the preferred interface to yield on-hardware demonstrations for robots to learn from~\cite{billard2016learning}.

\section{System Design}

We designed a pair of chopsticks for motion-capture (``MoChop") and an interface for users to teleoperate a robot holding chopsticks (``TeleChop"). Both were adapted from normal chopsticks (``Chop"). See Fig. \ref{fig:system_hardware}. All methods used the same consumer-grade titanium chopsticks that differ only in colors. 

\subsection{MoChop}
We 3D printed five light-weight, ball-shaped markers, wrapped them in reflective material and mounted them onto MoChop. To ensure users could still hold the chopsticks, we placed the markers near the tips and tails of chopsticks at different positions on each stick. Note the markers changed the weight distribution and added constraints to finger positions, e.g. users cannot hold the chopsticks at the tail.

To track the motion of MoChop held by users, we used the OptiTrack motion capture system \cite{optitrack} with 11 cameras (Fig. \ref{fig:system_mocapcage}). The system uses optical reflection to track the position of markers. Its tracking error is up to $0.4$mm, and the tracking updates at $100$Hz. From the tracked markers' positions, we extracted MoChop's pose.

\subsection{TeleChop}

We custom built a 6-DOF robot manipulator, assembled from components provided by HEBI Robotics ~\cite{hebi}. We attached the TeleChop to this robot's end-effector. TeleChop has an actuated pair of chopsticks. The two chopsticks operated on the same plane: one was fixed at the actuator's body, and the other is attached at the actuator's output shaft. We used HEBI's X-Series actuators ~\cite{hebi} since they provide built-in controllers for position, velocity, or torque control, running at a loop rate of $1$KHz. 

\subsection{Teleoperation Interface}
To initiate teleoperation, users held MoChop to match TeleChop's orientation. During teleoperation, users moved MoChop (\emph{leader}) to teleoperate the robot mounted with TeleChop (\emph{follower}). This further constrains chopsticks' movement because of the motion-retargeting problem induced by the difference between the human arm and the robot arm. We designed a controller that guide TeleChop to smoothly mimic MoChop's pose. Depending on how the user held MoChop, the two chopsticks may be on different planes. We projected the two chopsticks in MoChop onto the same plane to ensure TeleChop could follow the projected pose. This became the target pose that our controller tracked. 

Our controller translated MoChop poses to joint commands for the robot. Upon receiving the desired pose, the controller first computed an Inverse Kinematics solution. However, since smooth chopstick trajectories for humans could require jumps in the robot's joint space, we applied a convolution smoother to get the joint position command, effectively enabling tremor cancelling. We used a position PID controller to follow the joint position command. To add smoothness to the robot's movement, we also supplied a torque command based on gravity compensation and passed it through a torque PID controller. \footnote{The amount of torque commanded was based on the mass of the robot as specified on the manufacturer's datasheet ~\cite{hebi}} We added both PID controllers' PWM outputs to compose the final command.

\begin{table*}[!htbp]
\vspace{1em}
\includegraphics[width=.95\linewidth]{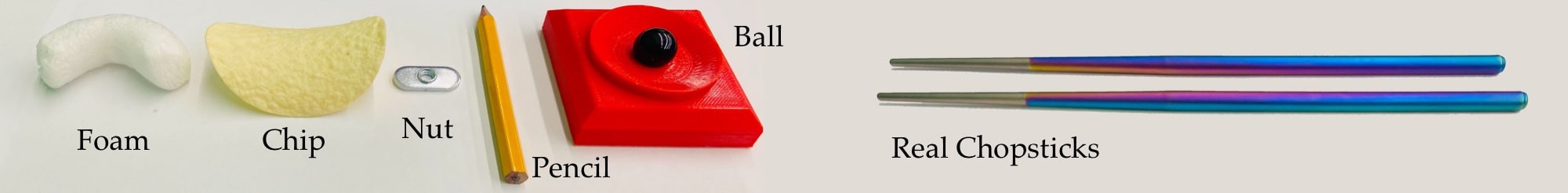}
\centering
\begin{tabular*}{.95\textwidth}{|c@{\extracolsep{\fill}}|ccccc|}

\hline
Name
    & Foam
    & Chip
    & Nut
    & Pencil
    & Ball \\
\hline
L $\times$ W $\times$ H (mm) \hspace{2pt} & 50$\times$25$\times$20& 65$\times$45$\times$15& 25$\times$10$\times$3& 86$\times$7$\times$6& 14$\times$14$\times$14\\
Texture & rough& rough& slippery & slippery & slippery      \\
Geometry & curved &curved &flat & long &spherical \\
Deformation & compliant  & brittle & hard  & hard & hard \\
\hline

\end{tabular*}
\caption{Different objects to pick up.}
\label{table:objects}
\end{table*}

\section{Experiments}
We conducted an in-lab user study with human subjects in which participants performed a series of pick-and-place manipulation tasks using chopsticks on a variety of objects. These experiments aimed to compare human performance on pick-and-place manipulation tasks using TeleChop, MoChop and Chop. We also wanted to explore how humans adapted to different interfaces during this process. Our object set (foam, chip, nut, pencil and ball) for the manipulation tasks included objects with varying levels of chopstick grasping difficulty. See Table~\ref{table:objects}. The user study was conducted in accordance with our University’s Institutional Review Board (IRB) review.

\subsection{Participants}
We recruited 25 human participants (13 male, 11 female, 1 non-binary of age $M = 27.28$, $SD = 7.89$) for our human-subject studies. The participants had various amount of experience using chopsticks ($M=14.44$, $SD = 8.79$ years of experience). Fourteen out of 25 subjects reported having experience with teleoperation, but none were previously exposed to our system. To offset individual differences among subjects, we chose a within-subjects design, where all subjects performed the same set of grasping tasks under the same conditions with different orders of tasks.

\subsection{Experiment Procedure}
Before beginning the experiments, participant signed a consent form and reported demographic information (procedure approved by the IRB review of the University of Washington). They were informed that the experiment was intended to evaluate their interactions with all three grasping methods. Prior to the recorded trial, participants went through a training procedure. First they held the Chop and then the MoChop, trying to open and close them. They watched the researcher demonstrate how to use TeleChop and then tried to initiate the teleoperation by matching the orientation of MoChop and TeleChop. During training, subjects interacted with Chop, MoChop and Teleop but not with any object in the recorded trials. The subjects finished training by picking up a piece of broken foam \emph{just once} using TeleChop.

Subjects then proceeded to the formal trials to be recorded. All subjects tried to pick up \emph{five} different items using all \emph{three} methods for grasping. Participants manipulated Chop and MoChop to directly pick up items; for TeleChop, they held MoChop to teleoperate the robot, who was holding TeleChop, to pick up items. For each combination of item and method, they had \emph{three} trials. Each subject therefore contributed 45 trials in total (5 objects $\times$ 3 methods $\times$ 3 trials). 

During each trial, participants were asked to pick up a specified object and hold it statically in the air for 1 second to show the grasp was firm. We defined success strictly as procuring the object \emph{in the first attempt}. If the chopsticks moved the object without procuring it, or the object immediately slipped away from chopsticks after being picked up, the trial failed. But the subjects were allowed to re-try the task until procuring the object or 20 seconds elapsed.
We chose 20 seconds as the maximum trial length because we wanted to (1) give subjects an opportunity to learn from interacting with the object, and (2) to control the total trial time so the subject would be less likely to feel tired or frustrated. 

Each subject worked with a randomized order of objects. For each object, the subject used all three methods. They completed all methods for one object before moving on to another. The subject might have gained more experience dealing with the object by the final method. We therefore randomized the order of methods for each object per person to ensure that each method’s data is not skewed. Upon finishing all tasks for one object, the subject rated the difficulty and comfort of each method on a 5-point Likert scale.

Upon completing all trials, participants responded to an open-ended post-task questionnaire. Samples of all questionnaires (pre-task, during-task, post-task) are available in \cite{questions}.

\subsection{Data Acquisition}
We recorded all trials using two RGB cameras and collected written questionnaires from subjects. We tracked and recorded MoChop movements during both MoChop's and TeleChop's trials. We recorded joint commands and robot states during teleoperation trials.

\begin{figure*}[!htbp]
\vspace{1em}
\centering
  \includegraphics[width=0.98\linewidth]{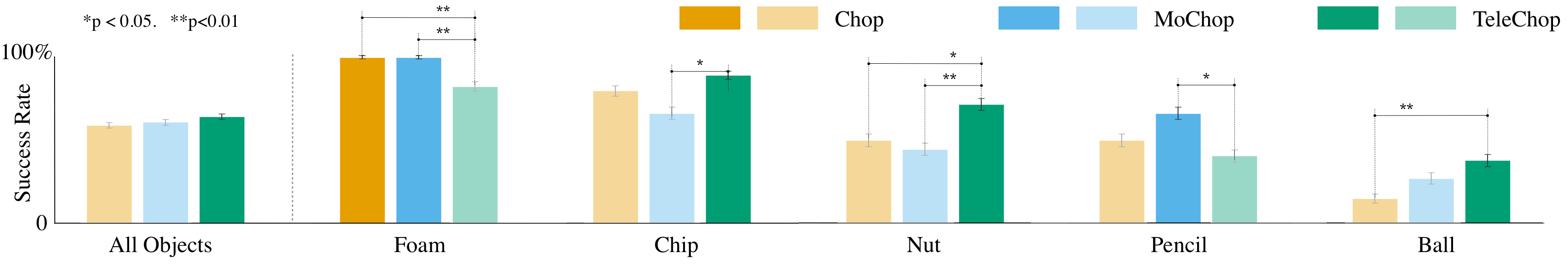}
  \caption{Success rate for each method. Error bars indicate 95\% confidence interval. No significant variance was observed across the three methods for all objects ($p=0.44$). However, for each individual object, we observed significant variance across methods ($p<0.05$ in ANOVA). The most successful method for each object is highlighted with high saturation color. 
  }
  \label{fig:by_method_and_objects}
\end{figure*}

\section{Analysis}

We evaluated how participants' performance varied by 1) object, 2) control interface and 3) chopstick expertise. We studied how participants adjusted their manipulation strategies for different objects, methods and after failures as a proxy of the learning effects. We also compared the subjective ratings with the objective success rate for each method and object.

\subsection{What factors affected the success of grasping?}

\paragraph{\textbf{Subjective rating of object difficulty matched objective success rate}}

\begin{figure}[!htbp]
  \centering
  \includegraphics[width=.95\linewidth]{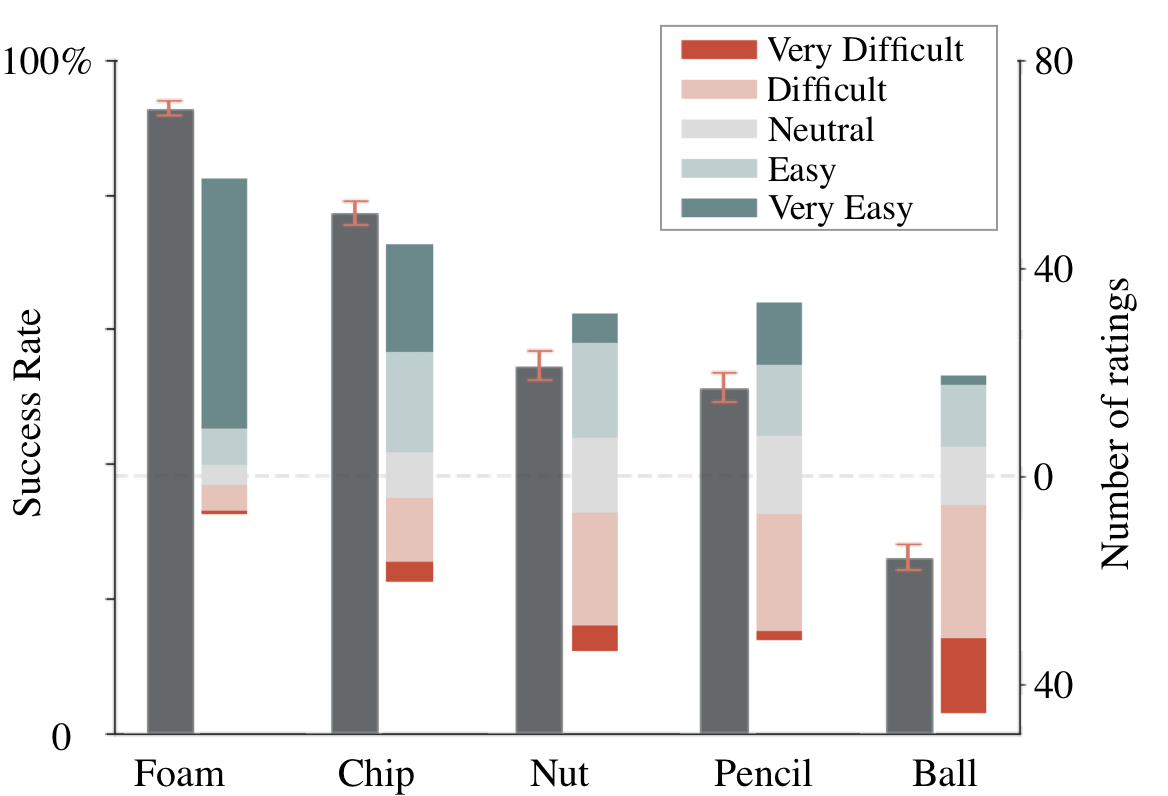}
  \caption{Success rate (gray bar) and subjective rating (multicolor bars) of difficulty for each object. From left to right, objects were easy to difficult.}
  \label{fig:object_success_and_difficulty}
\end{figure}

In Fig. \ref{fig:object_success_and_difficulty}, the variation in participants' success rates shows that the selected set of tasks presents a full spectrum variance of difficulty (ANOVA F=$38.49$, $p< 0.001$), ranging from very easy (foam) to very difficult (ball). The ranking derived from subjective ratings of difficulty roughly correlates with the corresponding ranking of performance, with the only exception being the order between the nut and the pencil. Paired-T tests on subjective ratings across objects were all significant ($p < 0.0001$). Paired-T test on success rates across objects were all significant ($p < 0.01$) except between nut and pencil ($p=0.65$).

\paragraph{\textbf{Different grasping methods affected the success rate for each object}} Fig. \ref{fig:by_method_and_objects} depicts the success rate per method and how performance varied per method for each object. Overall performance was similar (F=$0.8256$, p=0.44) between the grasping methods, although TeleChop performed slightly better.

However, for each individual object, we observed statistically significant differences across methods  ($F>3$, $p<0.04$). Chop and MoChop were significantly better ($p < 0.005$) at picking up foam, while TeleChop was significantly better at picking up a nut ($p < 0.05$). Foam was the simplest task and participants had 100\% success rate when directly using Chop and MoChop. They were probably familiar with the environment, and more specifically, the interaction between chopsticks and objects; unfamiliarity with teleoperation might explain the drop in success rate for this object. A nut, on the other hand, is a flat and slippery item that required a firm grasp after being picked up. Failures occurred mostly because the nut slipped from chopsticks after being lifted up, possibly because users were tired of supplying a concentrated force on their fingers. Teleoperation could alleviate this problem by offering a firm grasp without participants supplying force. More analysis on TeleChop's force output is in Sec.~\ref{subsec:position_proxy_force}.

\paragraph{\textbf{Participants improved performance within $3$ trials}}

\begin{figure*}[!htbp]
\vspace{1em}
\centering
  \includegraphics[width=0.98\linewidth]{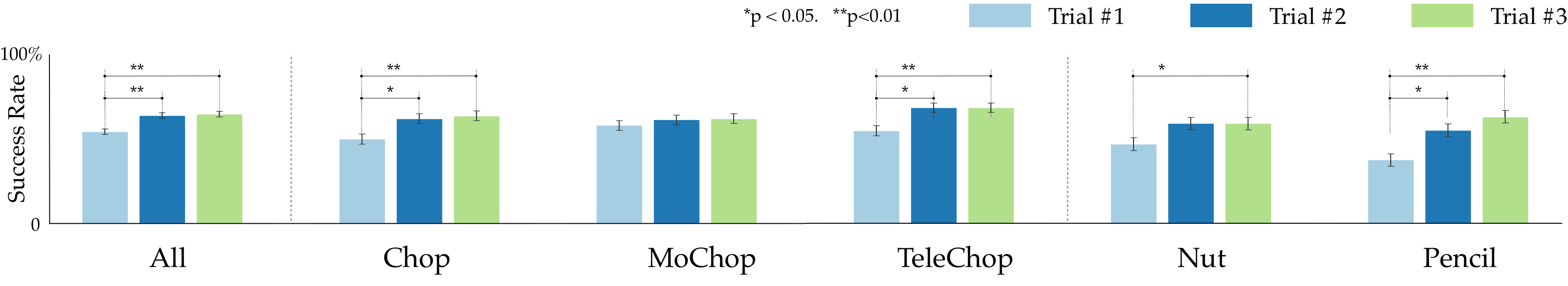}
  \caption{Success rate over trials}
  \label{fig:learning}
\end{figure*}

How quickly did subjects learn to adapt to the use of chopsticks for these trials? We looked at the change in success rate over the three trials as a proxy for estimating the effect of learning on participants' performance. For each task (5 in total) and each method (3 in total), each subject attempted 3 trials. Fig. \ref{fig:learning} depicts the success rate per trial number. As the trial count increased, performance increased, suggesting the existence of a learning effect (F=5.47, p$<0.001$). The variance in performance across trials was significant for Chop and TeleChop methods ($F=3.79$, $p = 0.027$, and $F=3.6$, $p = 0.032$ respectively), suggesting that subjects had an easier time adjusting to these two methods through trials.
Additionally, subjects significantly increased performance over trials when picking up a nut and pencil. For the nut, they might have realized the need to supply a firm grasp after failed trials. For the pencil, subjects might have mastered a more effective grasping point that was closer to CoM after experiencing failures. Further qualitative examination of how participants adjust their manipulation strategies is in Sec.~\ref{subsection:adjust_strategy}.

\paragraph{\textbf{Chop expertise doesn't imply TeleChop expertise}}

\begin{figure}[!htbp]
  \centering
  \includegraphics[width=.9\linewidth]{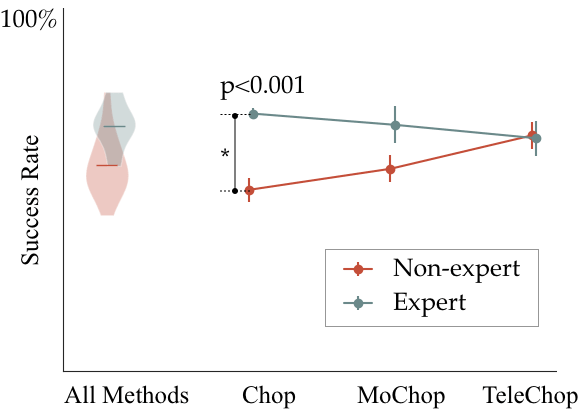}
  \caption{Success rate and chopstick expertise.}
  \label{fig:experts}
\end{figure}

We identified expert subjects by choosing individuals that achieved higher than median success rates when using Chop to pick up items ($\mathcal{SR}_{\text{expert}} > $ 60\%).  Based on this standard, our sample consisted of 11 expert and 14 non-expert subjects. Fig. \ref{fig:experts} depicts the performance of expert and non-expert chopstick subjects. Both cohorts had statistically significantly different success rates when using \emph{Chop} ($p<0.001$) but not when using MoChop or TeleChop ($p=0.06, 0.89$ respectively). Non-experts had a statistically significant improvement using TeleChop relative to using Chop ($p<0.01$). We suspect that non-expert participants may have had less stable and precise control of chopsticks, which would be critical in directly picking up a nut. However, teleoperation removed this requirement and therefore enabled non-experts to perform better. This suggests that teleoperation can be a desirable interface for future data collection, especially for collecting data on certain \emph{challenging tasks for humans}, such as prolonged grasping, as researchers can improve both hardware and controllers to alleviate the burden of controls from users.

\paragraph{\textbf{Given chances to re-try after a failed grasp, participant success rates can increased to above 75\% even for the most difficult task}}

\begin{figure}[h!]
  \includegraphics[width=0.95\linewidth]{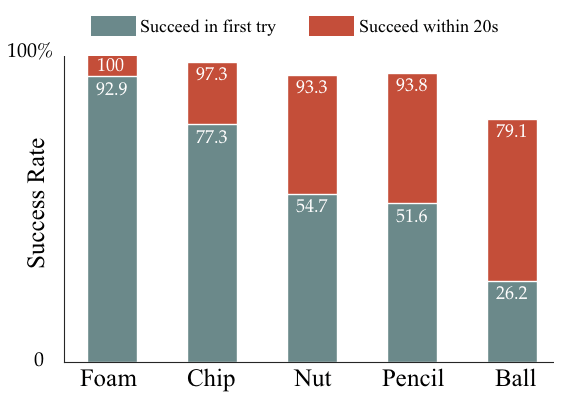}
  \caption{Given chances to re-try, subjects achieved greater success within 20 seconds. }
  \label{fig:repair}
\end{figure}

Many robotics grasping tasks evaluate success based on one-shot grasping, i.e., whether the robot successfully grasps the object in one try. However, humans learn from and adapt to failures ~\cite{bhattacharjee2019towards}.
In our experiment, we let subjects re-try a task after a failed trial, even though the first failed attempt might have changed the object's configuration. We evaluated re-try success rates based on whether subjects could pick up the same  object within 20 seconds (see Fig. \ref{fig:repair}.) We see a significant boost in the success rate where, within 20 seconds, subjects managed to pick up the most difficult item (glass ball) with a 79.1\% success rate even though their initial success rate at first try was only 26.2\%. Humans demonstrated an impressive robustness, which suggest that alternative metrics for evaluating manipulation performance might consider allowing successive tries. 

\subsection{How did participants adjust their manipulation strategies?}
\label{subsection:adjust_strategy}
\begin{figure*}[t!]
\vspace{1em}
\centering
\begin{subfigure}{.28\textwidth}
  \centering
  \includegraphics[width=.46\linewidth]{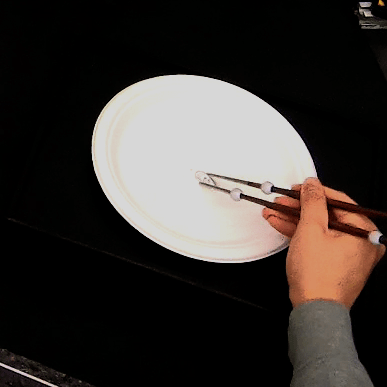}
  \includegraphics[width=.46\linewidth]{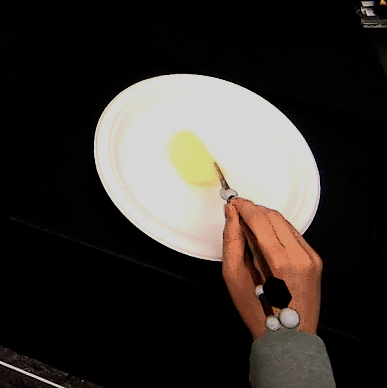}
  \caption{Nut versus Chip}
  \label{fig:strategy_nut_vs_chip}
\end{subfigure}
\begin{subfigure}{.28\textwidth}
  \centering
  \includegraphics[width=.46\linewidth]{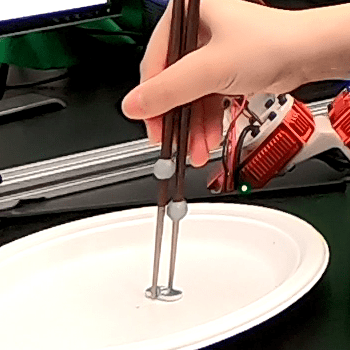}
  \includegraphics[width=.46\linewidth]{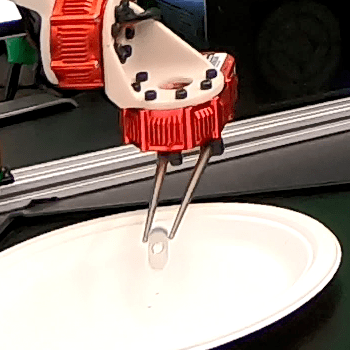}
  \caption{MoChop versus TeleChop}
  \label{fig:strategy_mochop_vs_teleop}
\end{subfigure}
\begin{subfigure}{.42\textwidth}
  \centering
  \includegraphics[width=.31\linewidth]{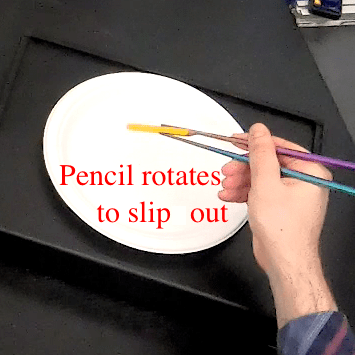}
  \includegraphics[width=.31\linewidth]{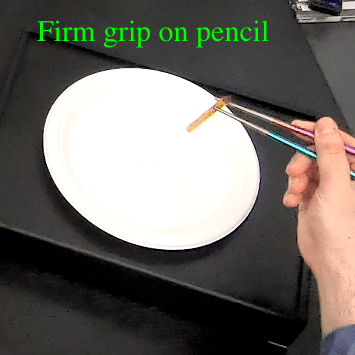}
  \includegraphics[width=.31\linewidth]{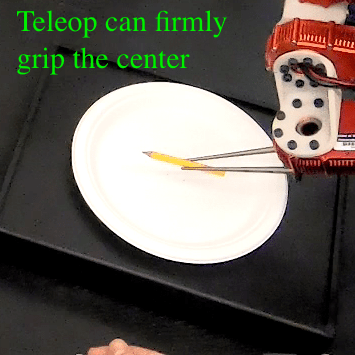}
  \caption{Adapting to failures}
  \label{fig:strategy_adapt}
\end{subfigure}
\caption{Subjects adjusted their manipulation strategies conditioned on (a) objects, (b) methods, and (c) adapting to failures. Each set of pictures shows the same user.}
\label{fig:adapt}
\end{figure*}

\paragraph{\textbf{Subjects applied different strategies for different shapes of objects}} For a flat object like nut, almost all subjects held the chopsticks parallel to the nut from the top view, as shown in Fig. \ref{fig:strategy_nut_vs_chip}. To adapt to the curved surface of chips, 21 of 25 subjects rotated the chopsticks to procure the object. Some subjects changed the rotation of the chopsticks several times \emph{without touching the object}, as if trying to find the optimal approach angle through visualizing the alignment. Visual information could play an important role in this adaptation. 


\paragraph{\textbf{For different methods, participants used different strategies for the same object}} Fig. \ref{fig:strategy_mochop_vs_teleop} shows a subject using MoChop to pick up the nut from a different angle than was used via TeleChop. Intuitively, holding chopsticks from the vertical angle could allow subjects to supply the strong force needed to pick up the Nut, whereas TeleChop could output the required force from any angle. We observed similar phenomena for chips: many participants used TeleChop to grip the chip \emph{on one pressure point} and used MoChop and Chop to \emph{lock} the chips between chopsticks. And for the glass ball: subjects needed to first grip the ball and then lift it up while retaining grasp strength. However, the glass ball frequently slipped during lifting. When using TeleChop, subjects seemed to pay more attention to gripping the ball and less to keeping the grip firm. These different strategies might also contribute to the performance difference.


\paragraph{\textbf{Participants changed strategies after failure}} As shown in Fig. \ref{fig:strategy_adapt}, the long pencil shape made it tricky to pick up. Most subjects tried to grip its center. Because some did not estimate the CoM accurately, the pencil rotated and dropped. After failure, many subjects adjusted the gripping point on the pencil and the contact point on the chopsticks. In this process, subjects also slowed down the manipulation motion, perhaps having realized that a rushed pick-up attempt made the pencil prone to rotate and drop. Interestingly, some had a different adaptation strategy when using TeleChop: 23 out of 26 subjects gripped the pencil at its geometry center, adjusted how much to close the leader chopsticks (and how much force TeleChop to output) until the robot could grasp the pencil firmly. 

The round slippery glass ball is among the most challenging object for chopsticks to pick up. Conceptually, picking up the ball is straight-forward: one need to close the chopsticks around the diameter of the ball. But placing the chopsticks exactly across the diameter of the ball was challenging. Many subjects went through a trial-and-error process, slightly lifting up the chopsticks and closing them around the ball, observing whether the ball was being picked up and adjusting the next grip accordingly. It allowed subjects to succeed at picking up the ball eventually, achieving an astonishing success rate of 79\% when allowed re-try. 



\subsection{Subjective Ratings}

Participants rated teleoperation as the least comfortable and the most difficult to use ($p<0.0001$), except when to pick up the ball. Participants explained that the negative ratings came from the sense of indirection added by the teleoperation interface, lack of haptic feedback and the misalignment between the robot arm and human arm. We also observed that subjects took significantly longer time on TeleChop's trial than Chop and MoChop.

However, participants' performance using teleoperation was empirically better than using other methods to pick up 3 out of the 5 objects chosen (chips, nut and ball shown in Fig.~\ref{fig:by_method_and_objects}). Possible reasons of such a contradiction include (1) teleoperation alleviated the burden of firm and stable control from the subject for certain tasks, (2) subjects found a strategy for teleoperation method that could achieve higher success rate but required more effort, and (3) subjects, \emph{feeling} that the teleoperation interface was more difficult and less comfortable, \emph{spend more effort and concentration} using this method. We found qualitative evidence supporting these hypotheses: (1) 15 of 25 subjects in the post-task questionnaire reported that TeleChop simplified grasping because it ``(was) easier to maintain a constant grip," ``(required) less effort on my hand to grip the object between the chopsticks," and ``(users) only need to care about the motion of chopsticks without, considering how much force to exert" , (2) 19 of 25 subjects developed a different strategy for TeleChop compared to the other two methods for grasping the same object, and (3) some subjects commented on how \emph{they} adapted to work with the TeleChop: ``I started focusing on the robot joints and how to move my arm in a way that translated to the robot joint, the placement tasks became easier." and ``It takes a little time to learn and be familiar with the movement (of TeleChop)". Therefore, the subjects adaptation to the robot and TeleChop's stable force output enable the teleoperation to achieve higher success rate on certain tasks but made it uncomfortable regardless. 


\subsection{Position as a proxy for force}
\label{subsec:position_proxy_force}
\begin{figure*}[!htbp]
\vspace{1em}
\centering
\begin{subfigure}{.19\textwidth}
  \centering
  \includegraphics[width=\linewidth]{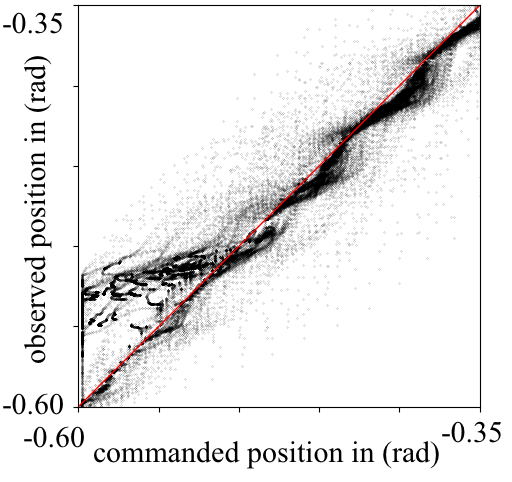}
  \caption{Foam}
  \label{fig:force_foam}
\end{subfigure}
\begin{subfigure}{.19\textwidth}
  \centering
  \includegraphics[width=\linewidth]{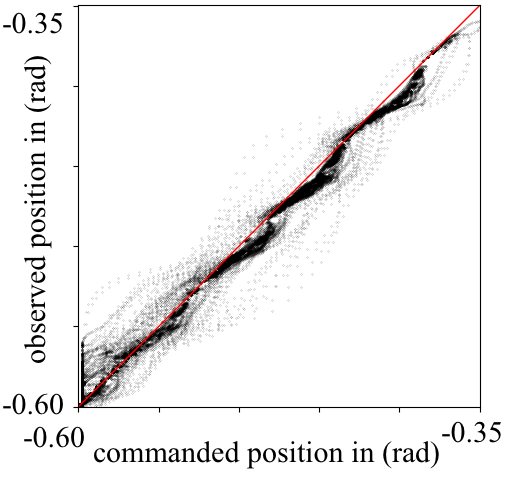}
  \caption{Chips}
  \label{fig:force_chip}
\end{subfigure}
\begin{subfigure}{.19\textwidth}
  \centering
  \includegraphics[width=\linewidth]{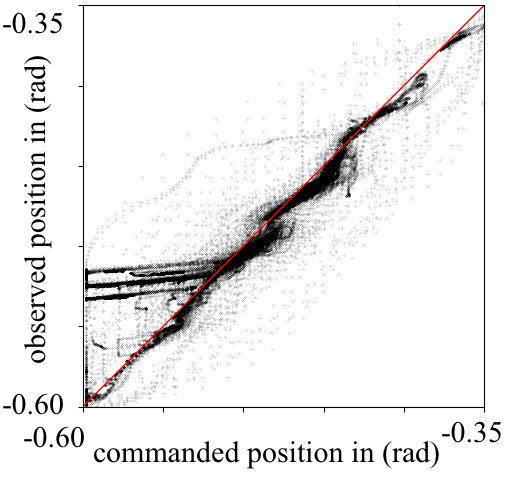}
  \caption{Nut}
  \label{fig:force_metal}
\end{subfigure}
\begin{subfigure}{.19\textwidth}
  \centering
  \includegraphics[width=\linewidth]{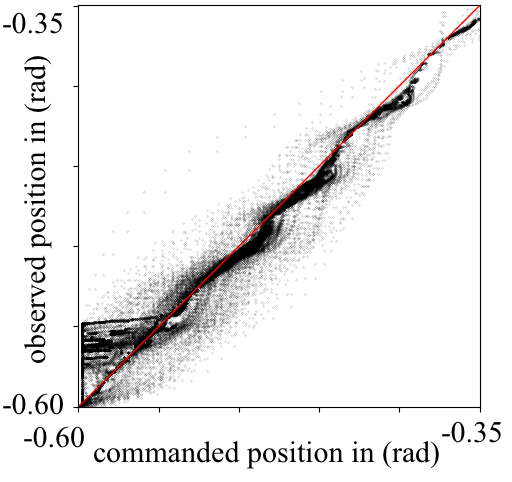}
  \caption{Pencil}
  \label{fig:force_pencil}
\end{subfigure}
\begin{subfigure}{.19\textwidth}
  \centering
  \includegraphics[width=\linewidth]{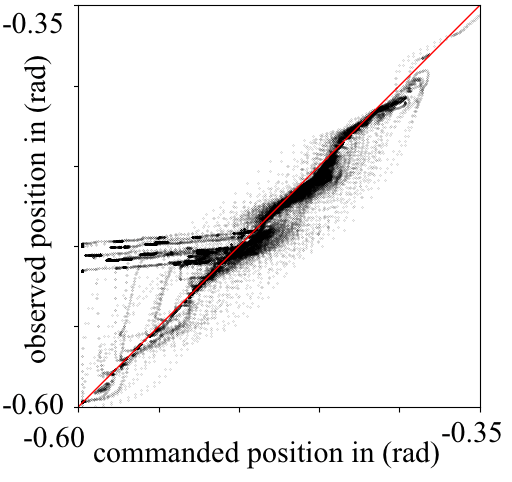}
  \caption{Ball}
  \label{fig:force_ball}
\end{subfigure}
\caption{Displacement between commanded position and observed position. The X-axis shows commanded position. The Y-axis shows observed position. (unit rads). A smaller number means closing chopsticks. The displacement is proportional to the torque output between chospticks.}
\label{fig:force}
\end{figure*}

One main benefit of our teleoperation interface is its ability to collect force commands without needing specialized force measurement devices. We achieved this by using position as a proxy for force. Subjects could close the leader chopsticks to direct TeleChop to close. However, TeleChop could hold a ball between its tips rather than closing the chopsticks as instructed. The error between the commanded and the actual position incurred a proportional torque output from our PID controller. To pick up any object that requires gripping, this mechanism is necessary. Fig. \ref{fig:force} shows the recordings of displacements, which indicate the generation of forces. Note that the clear displacement gaps for the Nut and Ball correspond to subjects gradually closing leader chopsticks to add force to firmly grip those objects. On post-task questionnaires, subjects commented on how firm the TeleChop grasp and that ``(they) didn't have to squeeze that hard to increase the tension." This might explain why TeleChop achieved highest success rate on the Nut and Ball. It also suggested that teleopeartion could be more helpful in picking up hard, small, and heavy pieces.

\section{Discussion}

\subsection{What interface is the best for collecting demonstrations?}

MoChop and Chop were better at picking up the foam and the pencil whereas TeleChop achieved the highest success rate picking up the chip, nut, and ball. Although TeleChop might be the most unintuitive and uncomfortable to use, subjects in our study demonstrated impressive flexibility and used teleoperation to achieve, overall, a comparable success rate to the other methods.

TeleChop has proved to be more helpful in picking up small, hard and slippery items that demanded precision and stability in control than soft, compliant, lightweight objects. This might be related to motor noise in human movement control. \cite{jones2002sources} proposed that higher muscle force generally comes with higher force variability. For tasks requiring strong and stable force output, e.g., surgical operation, teleoperation might allow for lower muscle forces, avoid muscle tremors, and could be the preferred method.

The subjective feedback suggested that the teleoperation interface tire out users at a faster rate than other methods, raising the question of whether the teleoperation interface can collect large amounts of demonstrations. Collecting data via motion-tracking could be less tiring and can perhaps capture more a realistic reflection of how humans use chopsticks. One possible remedy for this is to collect data in short-burst batches.  

However, an advantage of using teleoperation is that recorded trajectories can be replayed on the robot to accomplish the task. Transforming and replaying motion-captured data is similar to open-loop control, whereas replaying teleoperation data is closer to closed-loop control as the human demonstrator serves as the closed-loop controller for the robot during recording. This feature can be critical for approaches like learning from demonstrations. 

All things considered, we would recommend using a teleoperation interface to collect demonstrations for robotics manipulation with chopsticks.
Teleoperation superiority on certain tasks---even though it takes away haptic feedback and introduces additional delays---implies that teleoperation could boost human performance, e.g., offering tremor canceling in robotic surgery, supplying stronger force output in exoskeletons, or aiding people with disabilities. 

\subsection{Can we learn from human adaptation to a robotic arm?}

Human subjects have demonstrated an impressive capability to serve as the controller of imperfect hardware that is not precise to control, counters their intuition during usage, and lacks the haptic feedback that they are familiar with in challenging manipulation tasks. 

During human study, we observed multiple learning effects happening. The subject learned about different chopsticks methods, the robot interface, the object dynamics, and the manipulation strategies they used. We highlight how non-expert subjects using TeleChop could achieve comparable success rates to expert subjects using Chop. We quantify the subjects' improvement by studying the change in success rate over three trials, noting a marked increase in success. More importantly, humans recognize their mistakes immediately and alter their manipulation strategies during the course of a single task, achieving a median success rate of over 93\% but not on the first try. Therefore, recognizing the short-term learning factors in humans and formalizing them in algorithms may boost robotics manipulator's robustness. 

\subsection{Limitations}

Our teleoperation interface allows successful demonstrations of chopsticks manipulation, but it can be counter-intuitive and uncomfortable to use. It would be more beneficial to design the teleoperation robot to use a parallel configuration to human arms, have more stable movement, and enable more advanced tremor canceling. We used a simple controller and haven't tuned our system to completely eliminate the control noise. A smoother and more precise teleoperation system could improve the human experience and reduce the cognitive load. 

Nevertheless, our work brings out the potential of involving chopsticks in robotic manipulation and demonstrates how a teleoperated robot with chopsticks can pick-up challenging items without relying on a complicated end-effector. We intend to extend this work by learning from the demonstrations we collected and building a skillful robot at using chopsticks autonomously.




%



\section*{Acknowledgement}
This work was (partially) funded by the National Institute of Health R01 (\#R01EB019335), National Science Foundation CPS (\#1544797), National Science Foundation NRI (\#1637748), the Office of Naval Research, the RCTA, Amazon, and Honda Research Institute USA.


\bibliographystyle{IEEEtran}
\bibliography{main}

\end{document}